\documentclass[preprint,12pt]{elsarticle}

\usepackage{amssymb}

\usepackage{amsthm}
\newtheorem{mydef}{Definition}

\usepackage{balance}
\usepackage{booktabs}
\usepackage{amsmath}
\usepackage{tabularx}
\usepackage{subfigure}
\usepackage{graphicx}
\usepackage{multirow}
\usepackage{rotating}
\usepackage{algorithmic}

\journal{Neurocomputing}

\begin{document}

\begin{frontmatter}

\title{Combining Multiple Clusterings via Crowd Agreement Estimation and Multi-Granularity Link Analysis}

\cortext[*]{Corresponding author. Present address: School of Information Science and Technology, Sun Yat-sen University, Guangzhou Higher Education Mega Center, Panyu District, Guangzhou, Guangdong, 510006, P. R. China. Tel.: +86-13168313819. Fax: +86-20-84110175.}
\author[1,4]{Dong Huang}
\ead{huangdonghere@gmail.com}
\author[1]{Jian-Huang Lai\corref{*}}
\ead{stsljh@mail.sysu.edu.cn}
\author[2,3]{Chang-Dong Wang}
\ead{changdongwang@hotmail.com}
\address[1]{School of Information Science and Technology,Sun Yat-sen University,Guangzhou,China}
\address[2]{School of Mobile Information Engineering, Sun Yat-sen University, Guangzhou, China}
\address[3]{SYSU-CMU Shunde International Joint Research Institute (JRI), Shunde, China}
\address[4]{Guangdong Key Laboratory of Information Security Technology, Guangzhou, China}

\begin{abstract}

The clustering ensemble technique aims to combine multiple clusterings into a probably better and more robust clustering and has been receiving an increasing attention in recent years. There are mainly two aspects of limitations in the existing clustering ensemble approaches. Firstly, many approaches lack the ability to weight the base clusterings without access to the original data and can be affected significantly by the low-quality, or even ill clusterings. Secondly, they generally focus on the instance level or cluster level in the ensemble system and fail to integrate multi-granularity cues into a unified model. To address these two limitations, this paper proposes to solve the clustering ensemble problem via crowd agreement estimation and multi-granularity link analysis. We present the normalized crowd agreement index (NCAI) to evaluate the quality of base clusterings in an unsupervised manner and thus weight the base clusterings in accordance with their clustering validity. To explore the relationship between clusters, the source aware connected triple (SACT) similarity is introduced with regard to their common neighbors and the source reliability. Based on NCAI and multi-granularity information collected among base clusterings, clusters, and data instances, we further propose two novel consensus functions, termed weighted evidence accumulation clustering (WEAC) and graph partitioning with multi-granularity link analysis (GP-MGLA) respectively. The experiments are conducted on eight real-world datasets. The experimental results demonstrate the effectiveness and robustness of the proposed methods.

\end{abstract}


\end{frontmatter}


\section{Introduction}
\label{sec:intro}
Data clustering is a fundamental and very challenging problem in data mining and machine learning. The purpose is to partition unlabeled data into homogeneous groups, each referred to as a cluster. Data clustering requires a distance metric for evaluating the similarity between data instances, which, without prior knowledge of cluster shapes, is hard to specify. In the past few decades, a large number of clustering algorithms have been developed \cite{xu93_rpcl,li07_hmac,zhang09_ai,zhao10_nc,ebcl11,li11_cvpr,meap13,svstream13,psvdd13}. However, there is no single clustering method which is able to identify all sorts of cluster shapes and structures in data.

For the same dataset, different methods, or even the same method with different initializations or parameter settings, may lead to very different clustering results. It is extremely difficult to decide which method would be the \emph{proper} one for a given clustering task, not to say how to properly specify the initialization and parameter setting for the chosen method. Each method has its own merits as well as weaknesses. Different clusterings generated by different methods or with varying parameters can provide multiple views of the data. How to combine the information of different clustering results for obtaining a better and more robust clustering remains a very challenging problem \cite{jain10_survey,vega_pons11_survey}.

In recent years, many clustering ensemble approaches have been developed, which aim to combine multiple clusterings into a probably better and more robust clustering by utilizing various techniques \cite{strehl02,fern04_bipartite,Fred05_EAC,topchy05,diversity06,li07,iamon08_icds,Domeniconi09,wang09_pr,Mimaroglu11_pr,iam_on11_linkbased,yi_icdm12,franek13_pr}. However, in most of the existing methods, there are mainly two aspects of limitations. Firstly, many of the clustering ensemble approaches lack the ability to weight the base clusterings without access to the original data features, which makes them vulnerable to low-quality clusterings and probable to be affected significantly by low-quality clusterings (or even ill clusterings). Secondly, they mainly focus on the instance level or the cluster level in the ensemble system and fail to fuse multi-granularity information into a unified model. In order to address these two limitations, in this paper, we propose a clustering ensemble framework based on crowd agreement estimation and multi-granularity link analysis. By exploring the relationship among the base clusterings, we present a novel clustering validity measure termed normalized crowd agreement index (NCAI), which is able to evaluate the quality of base clusterings in an unsupervised manner and provides information for treating each base clustering accordingly. The source aware connected triple (SACT) similarity is introduced for analyzing the similarity between clusters with regard to their common neighbors and source reliability. Besides the relations between base clusterings and between clusters, we further investigate the linkage between data instances and clusters and incorporate the information from the three levels of granularity in a unified framework. In our previous work \cite{huang_iscide13}, we introduced the consensus function termed graph partitioning with multi-granularity link analysis (GP-MGLA). This paper is a major extension of our previous work on clustering ensemble. In this paper, more comprehensive literature and motivation are provided. Besides that, we propose another novel consensus function termed weak evidence accumlation clustering (WEAC), which is developed from the conventional evidence accumulation clustering (EAC) \cite{Fred05_EAC} and capable of dealing with ill clusterings by incorporating the clustering validity cue into the ensemble process. Extensive experiments are further conducted on real-world datasets for evaluating the proposed methods against several baseline clustering ensemble methods.

The remainder of this paper is organized as follows. In Section~\ref{sec:related_work}, we review the related work of the clustering ensemble technique. In Section~\ref{sec:ensemble_problem}, we describe the formulation of the clustering ensemble problem. In Section~\ref{sec:cae}, we present the crowd agreement estimation mechanism. The source aware connected triple (SACT) similarity is introduced in Section~\ref{sec:sact}. In Section~\ref{sec:ensemble_technique}, we propose two novel consensus functions termed weighted evidence accumulation clustering (WEAC) and graph partitioning with multi-granularity link analysis (GP-MGLA) respectively. The experimental results are reported in Section~\ref{sec:experiment}. We conclude this paper in Section~\ref{sec:conclusion}.

\section{Related Work}
\label{sec:related_work}

\begin{figure*}[!t]
\hskip 0.2in
\begin{center}
{
{\includegraphics[width=0.8\columnwidth]{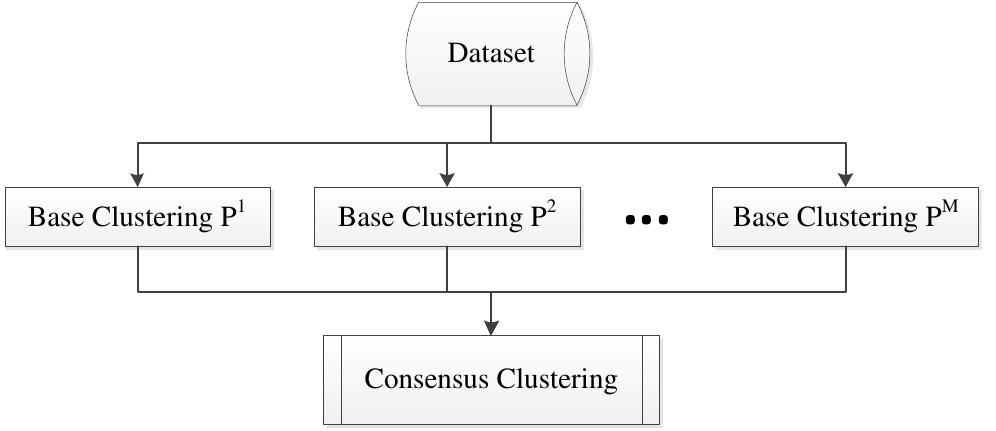}}}
\caption{The clustering ensemble process.}
\label{fig:flowchart}
\end{center}
\vskip -0.2in
\end{figure*}

Clustering ensemble is also known as clustering combination or clustering aggregation, which aims to combine multiple clusterings, each referred to as a base clustering (or an ensemble member), to obtain a so-called consensus clustering. As illustrated in Fig.~\ref{fig:flowchart}, the clustering ensemble process involves two steps: the first step is to generate multiple clusterings for a given dataset; and the second step is to construct the consensus clustering from the ensemble of base clusterings using different consensus functions.

Given a dataset, the ensemble of base clusterings can be generated by running different clustering algorithms \cite{Mimaroglu11_pr,yi_icdm12,huang_iscide13}, running the same algorithm with different initializations and parameters \cite{Fred05_EAC,iamon08_icds,wang09_pr,iam_on11_linkbased}, clustering via sub-sampling the data repeatedly \cite{strehl02,fern04_bipartite}, or clustering via projecting the data onto different subspaces \cite{strehl02,fern04_bipartite,topchy05,Domeniconi09}. Compared to generating base clusterings, how to combine multiple base clusterings, i.e., how to design the consensus function, is much more important and challenging in the clustering ensemble problem.

In the past few years, many consensus functions have been developed to fuse information from multiple clusterings \cite{strehl02,fern04_bipartite,Fred05_EAC,topchy05,diversity06,li07,iamon08_icds,Domeniconi09,wang09_pr,Mimaroglu11_pr,iam_on11_linkbased,yi_icdm12,franek13_pr}. These approaches can be classified into mainly three categories, namely, (i) the median partition based methods \cite{cristofor02,topchy05,franek13_pr}, (ii) the pair-wise co-occurrence based methods \cite{Fred05_EAC,li07,wang09_pr}, and (iii) the graph partitioning based methods \cite{strehl02, fern04_bipartite,Domeniconi09}.

In the median partition based approaches \cite{cristofor02,topchy05,franek13_pr}, the clustering ensemble problem is formulated into an optimization problem, aiming to find the partition/clustering that maximizes the similarity between the the partition and the base clusterings, over the space of all partitions. The median partition problem is NP-complete \cite{topchy05}. Instead of finding the optimal solution over the huge space of all possible partitions, Cristofor and Simovici \cite{cristofor02} used the genetic algorithm to obtain an approximative solution where the clusterings are represented by chromosomes. Topchy et al. \cite{topchy05} cast the median partition problem into a maximum likelihood problem, as a solution to which the consensus clustering is found using the EM algorithm. Franek and Jiang \cite{franek13_pr} reduced the median partition problem to the Euclidean median problem by clustering embedding in vector spaces and found the median vector by the Weiszfeld algorithm \cite{Weiszfeld09}. Then an inverse transformation would be performed to convert the median vector into a clustering, which was taken as the consensus clustering.

The pair-wise co-occurrence based approaches \cite{Fred05_EAC,li07,wang09_pr} construct the similarity between data instances by considering how many times they occur in the same cluster in the ensemble of base clusterings. Fred and Jain \cite{Fred05_EAC} introduced the evidence accumulation clustering (EAC) method, which used the co-association matrix to measure the similarity between instances. Then the hierarchical agglomerative clustering algorithms \cite{jain10_survey}, e.g., single-link (SL) and average-link (AL), can be performed on the co-association matrix and thus the consensus clustering is obtained. Li et al. \cite{li07} analyzed the co-association matrix and proposed a novel hierarchical clustering algorithm by utilizing the concept of normalized edges to measure the similarity between clusters. Wang et al. \cite{wang09_pr} generalized the EAC method and proposed the probability accumulation method, which took into consideration the sizes of clusters in the ensemble.

Another category of clustering ensemble is based on graph partitioning \cite{strehl02, fern04_bipartite,Domeniconi09}. Strehl and Ghosh \cite{strehl02} modeled the ensemble of clusterings in a hypergraph structure where the clusters are treated as hyperedges. For partitioning the graph and obtaining the consensus clustering, they further proposed three graph partitioning algorithms, namely, the cluster-based similarity partitioning algorithm (CSPA), the hypergraph-partitioning algorithm (HGPA), and the meta-clustering algorithm (MCLA). Fern and Brodley \cite{fern04_bipartite} formulated the clustering ensemble into a bipartite graph where both the data instances and clusters are represented as graph nodes. An edge between two nodes exists if and only if one of the nodes is a data instance and the other node is the cluster containing it. The consensus clustering is obtained by partitioning the graph into a certain number of disjoint sets of graph nodes.

Many of the existing clustering ensemble approaches implicitly assume that all the base clusterings contribute equally to the ensemble system and can be affected significantly by low-quality clusterings or even ill clusterings. In recent years, some efforts have been made to weight the base clusterings with regard to the clustering validity. Vega-Pons et al. \cite{vega_pons10} exploited several property validity indexes (PVIs), namely, Variance (VI), Connectivity (CI), Silhouette Width (SI) and Dunn index (DI), to assign a weight to each partition in the ensemble and proposed a new clustering ensemble method based on kernel functions. Vega-Pons et al. \cite{vega_pons_PRL11} also extended the conventional EAC method by weighting the partitions based on the PVIs. These PVIs need access to the original feature vectors, which are not supposed to be given for the consensus process in the formulation of this work as well as many other clustering ensemble frameworks \cite{fern04_bipartite,Fred05_EAC,li07,iamon08_icds,wang09_pr,Mimaroglu11_pr,iam_on11_linkbased,yi_icdm12,franek13_pr}. Li and Ding \cite{Li_WCC08} proposed the weighted consensus clustering (WCC) method, where the weights of the base clusterings are determined via an optimization process based on the nonnegative matrix factorization. The optimization process is computationally expensive when dealing with large datasets. Fern and Lin \cite{Fern08_selection} proposed a clustering ensemble selection framework which selects a subset of partitions from a large library of partitions. The ensemble selection process in \cite{Fern08_selection} can be viewed as weighting the partitions in the ensemble with either 1 or 0, where 1's indicate the preserved partitions and 0's indicate the deleted ones. However, the ensemble selection scheme lacks the flexibility of weighting the selected members in accordance to their quality.

\section{The Clustering Ensemble Problem}
\label{sec:ensemble_problem}

The purpose of a clustering algorithm is to discover the structure of clusters in a given dataset. The clustering result can be either a hard partition or a fuzzy partition for the dataset. The clustering ensemble technique aims to combine multiple partitions for achieving a better partition. In this paper, we focus on combining hard patitions of data.

Given a dataset $\mathcal{X}=\{x_1,x_2,\dots,x_n\}$, where $x_i$ is the $i$-th data instance and $n$ is the number of instances in $\mathcal{X}$. A partition (or clustering) of $\mathcal{X}$ is generated by running a clustering algorithm with some specific parameters. Each cluster in a partition consists a certain number of data instances. Different clusters in the same partition do not intersect with each other. And the union of all clusters in a partition covers the entire dataset.

Formally, let
\begin{equation}
\label{eq:partition}
P^i=\{C_1^i,C_2^i,\dots,C_{n_i}^i\}
\end{equation}
be a partition of $\mathcal{X}$, where $C_j^i$ denotes the $j$-th cluster and $n_i$ is number of clusters in $P^i$. Then we have $\forall C_j^i\in P_i, C_j^i \neq \emptyset$, $\forall j\neq k, C_j^i\cap C_k^i=\emptyset$, and $\cup_{j=1}^{n_i}C_j^i=P^i$.

In a clustering ensemble system, each partition is referred to as a base clustering. With the partitions generated by different algorithms or the same algorithm with different parameters and initializations, we can obtain the ensemble of $M$ base clusterings, which is denoted as
\begin{equation}
\label{eq:ensemble}
\mathcal{P}=\{P^1,P^2,\dots,P^M\},
\end{equation}
where $P^i$ represents the $i$-th base clustering in $\mathcal{P}$. For convenience, the set of all clusters in the ensemble is denoted as $\mathcal{C}=\{C_1,C_2,\dots,C_{n_c}\}$, where $C_i$ is the $i$-th cluster in $\mathcal{C}$. As is defined, it holds that $\mathcal{C} = \cup_{i=1}^M P^i$ and $n_c=\sum_{i=1}^M n_i$.

The multiple partitions of $\mathcal{X}$ provide multiple looks at the dataset. The problem is to use the information provided by the the ensemble of multiple partitions to obtain a final partition solution $P^*$, which is generally referred to as the consensus clustering.

\section{Crowd Agreement Estimation}
\label{sec:cae}

In the clustering ensemble system, the base clusterings can be generated using a wide variety of clustering algorithms. Due to the diversity of clustering algorithms and datasets, it is not guaranteed that every base clustering is well constructed. The low-quality clusterings, or even ill clusterings, may affect the consensus process significantly. There is a need to distinguish the poor clusterings from the good ones and treat the base clusterings with regard to their quality. The critical problem here is how to evaluate the quality of the base clusterings without knowing the ground-truth.

Some algorithms have been developed to estimate the clustering quality using different criteria \cite{wu04_validity,faceli09,li12_nips}. Wu and Chow \cite{wu04_validity} proposed a clustering validity index based on inter-cluster and intra-cluster density. Faceli et al. \cite{faceli09} used the overall deviation and the connectivity to assess the quality of a clustering. The overall deviation of a clustering measures the overall distances between data instances and their corresponding cluster centers. The connectivity measures how often neighboring instances are assigned to the same cluster. Li and Latecki \cite{li12_nips} utilized the average silhouette coefficient to evaluate the quality of a cluster. The silhouette coefficient of a data instance measures how similar that instance is to the instances in its own cluster compared to the instances in the other clusters, whereas the quality of a cluster is estimated by the average of the silhouette coefficients of the instances inside it. These evaluation methods are only applicable to numerical data and need access to the original data features, which are not supposed to be given in the problem formulation of many clustering ensemble approaches \cite{fern04_bipartite,Fred05_EAC,li07,iamon08_icds,wang09_pr,Mimaroglu11_pr,iam_on11_linkbased,yi_icdm12,franek13_pr}. Rather than utilizing the information of data distribution, in this paper, we view the clustering ensemble as a crowd of individuals and estimate the quality of each individual via consulting the other individuals in the clustering ensemble.

In social and economic science, ``the wisdom of the crowd'' is the process of taking into consideration the collective opinion of a crowd of individuals rather than a single expert \cite{Surowiecki04}. The ground-truth labeling of a dataset can be viewed as an expert. As the ground-truth is not supposed to be known in unsupervised frameworks, we estimate the quality of a base clustering by collecting information from the crowd of base clusterings. Each base clustering is compared with the other ones and the average opinion of the crowd of individuals is obtained for quality estimation.

\begin{mydef}
Let $\mathcal{P}$ be an ensemble of base clusterings and $P^i$ be the $i$-th base clustering in $\mathcal{P}$. The crowd agreement index (CAI) for $P^i$ is defined as
\begin{equation}
\mathit{CAI}(P^i)=\frac{1}{M-1}\sum_{P^j\in\mathcal{P}, i\neq j} \mathit{Sim}(P^i, P^j),
\end{equation}
where $\mathit{Sim}(P^i, P^j)$ denotes the similarity between the two base clusterings $P^i$ and $P^j$.
\end{mydef}

We denote the base clustering that gains the maximum agreement from the crowd as the reference member. Then the reliability of the base clusterings is estimated by comparing their crowd agreement with that of the reference member and the normalized version of crowd agreement index can be computed.

\begin{mydef}
\label{def:ncai}
The normalized crowd agreement index of $P^i$ is defined as
\begin{equation}
\label{eq:ncai}
\mathit{NCAI}(P^i)=\frac{\mathit{CAI}(P^i)}{\max_{P^j\in \mathcal{P}} \mathit{CAI}(P^j)}.
\end{equation}
\end{mydef}

The basic idea here is to estimate the quality of a base clustering by collecting opinion from a crowd of diverse individuals. According to Definition~\ref{def:ncai}, for $i=1,2,\dots, M$, it holds that $\mathit{NCAI}(P^i)\in [0,1]$. In this paper we use the normalized mutual information (NMI) \cite{strehl02} as the similarity measure $\mathit{Sim}(P^i, P^j)$. The greater the NCAI value of a base clustering is, the better its quality is supposed to be.

\section{Source Aware Connected Triple}
\label{sec:sact}

In this section, we investigate the relationship among the clusters in the ensemble and introduce the source aware connected triple (SACT) which is able to measure the similarity of two clusters with regard to their common neighbors and the source reliability.

\begin{mydef}
\label{def:neighbor}
Two clusters $C_i$ and $C_j$ are neighbors if and only if they share some common data instances, i.e., $C_i\cap C_j\neq \emptyset$.
\end{mydef}

Each cluster is a set of data instances. The Jaccard coefficient \cite{levan71_nature} is often used to measure the similarity between two clusters (or two sets), which is computed as follows:
\begin{equation}
\label{eq:jaccard}
J(C_i, C_j) = \frac{|C_i\cap C_j|}{|C_i\cup C_j|},
\end{equation}
where $C_i$ and $C_j$ are two clusters and $|S|$ denotes the cardinality of the set $S$. The Jaccard coefficient takes into consideration the sharing instances of two clusters to measure their similarity. Therefore the Jaccard coefficient of two clusters in the same base clustering is always zero. If two clusters intersect, then they are directly related. If two clusters do not intersect but they share a certain number of common neighbors, then they are also related. Iam-On et al. \cite{iam_on11_linkbased} utilized the information of common neighbors of two clusters to justify their similarity, where, however, the reliability of these neighbors was not considered.

Each base clustering can be viewed as a source of clusters. The overall quality of the clusters in a base clustering is correlated to the quality of the base clustering containing them. In this paper, we estimate the reliability of a cluster by considering the quality of the corresponding base clustering and propose the source aware connected triple (SACT) to measure the similarity of two clusters with regard to their common neighbors and the reliability of these neighbors.

\begin{mydef}
\label{def:sact}
The SACT coefficient between two clusters $C_i$ and $C_j$ w.r.t. a cluster $C_k$ is defined as
\begin{equation}
\mathit{SACT}_{ij}^k = I_{\mathit{NCAI}}(P(C_k))\cdot\min(J(C_i,C_k),J(C_j,C_k)),
\end{equation}
where $P(C_k)$ denotes the base clustering that contains $C_k$ and
\begin{equation}
\label{eq:i_ncai}
I_{\mathit{NCAI}}(P^l)=(\mathit{NCAI}(P^l))^{\beta}
\end{equation}
is the influence of the NCAI of the base clustering $P^l$.
\end{mydef}

According to Definitions~\ref{def:ncai} and \ref{def:sact}, for $l=1,2,\dots,M$, it holds that $I_{\mathit{NCAI}}(P^l)\in [0,1]$. The parameter $\beta>0$ in Eq.~(\ref{eq:i_ncai}) is a parameter to adjust the influence of the NCAI. A greater value of $\beta$ leads to a bigger influence of the NCAI, which means the difference of NCAI values between high-confidence partitions and low-confidence partitions is enlarged. When $\beta=0$, the influence of NCAI disappears for all base clusterings, i.e., $\forall P^l\in\mathcal{P}, I_{\mathit{NCAI}}(P^l)=0$.

\begin{mydef}
\label{def:sact_2}
The SACT coefficient between two clusters $C_i$ and $C_j$ w.r.t. all the clusters in the ensemble $\mathcal{C}$ is defined as
\begin{equation}
\label{eq:sact_2}
\mathit{SACT}_{ij}=\sum_{C_k\in\mathcal{C}}\mathit{SACT}_{ij}^k.
\end{equation}
\end{mydef}

By definition, if $C_k$ is not a common neighbor between $C_i$ and $C_j$, then $\mathit{SACT}_{ij}^k=0$. Thus the SACT coefficient between two clusters w.r.t. all the common neighbors is identical to that w.r.t. all the clusters in the ensemble and can be computed by Eq.~(\ref{eq:sact_2}).

\begin{mydef}
\label{def:sact_sim}
The SACT similarity between two clusters $C_i$ and $C_j$ is defined as
\begin{equation}
\label{eq:sact_sim}
\mathit{SIM}_{\mathit{SACT}}(C_i,C_j)=\begin{cases}
1, &\text{if}~i=j,\\
\frac{\mathit{SACT}_{ij}}{\max_{\forall C_x,C_y\in \mathcal{C}}\mathit{SACT}_{xy}}, &\text{otherwise.}
\end{cases}
\end{equation}
\end{mydef}

The SACT similarity is computed on the basis of the the SACT coefficient. The pair of clusters with the maximum SACT coefficient is adopted as the reference pair of clusters, whose SACT similarity is defined to be 1. The SACT similarity of the other pairs of clusters is computed by comparing their SACT coefficient to that of the reference pair (see Eq.~(\ref{eq:sact_sim})). The SACT similarity between a cluster and itself is set to 1.

\section{Consensus Functions}
\label{sec:ensemble_technique}

In this section, we introduce two novel consensus fuctions which utilize multi-granularity information of the ensemble and are able to deal with ill base clusterings. In the following, we will describe the weighted evidence accumulation clustering (WEAC) method in Section~\ref{sec:weac} and the graph partitioning with multi-granularity link analysis (GP-MGLA) method in Section~\ref{sec:gp_mgla}.

\subsection{Weighted Evidence Accumulation Clustering (WEAC)}
\label{sec:weac}

In a base clustering, each data instance is assigned to a specific cluster, whereas two instances are either in the same cluster or in two different clusters. Without access to the original features, the affinity between two data instances can be assessed by their co-occurrence information in the ensemble of base clusterings.

\begin{mydef}
\label{def:similarity_matrix}
Let $P^l$ be a base clustering in the clustering ensemble $\mathcal{P}$. Let $P^l(i)$ be the cluster label of the instance $i$ in $P^l$. The $n\times n$ similarity matrix $S^l$ for $P^l$ is computed as follows:
\begin{eqnarray}
\label{eq:similarity_matrix}
&&S^l_{ij}=\begin{cases}
1, &\text{if}~P^l(i)=P^l(j),\\
0, &\text{otherwise,}
\end{cases}\\
&&\text{for $i=1,\dots,n$, $j=1,\dots,n$.}\nonumber
\end{eqnarray}
\end{mydef}

For each base clustering, say, $P^l$, a similarity matrix $S^l$ is constructed. If instances $i$ and $j$ occur in the same cluster in $P^l$, then $S^l_{ij}=1$; otherwise $S^l_{ij}=0$. The similarity matrix contains the pair-wise co-occurrence information of the corresponding base clustering. In the conventional evidence accumulation clustering (EAC) method \cite{Fred05_EAC}, the association matrix $A$ is obtained by averaging the similarity matrices of all the base clustering, that is
\begin{equation}
\label{eq:eac}
A=\frac{1}{M}\sum_{l=1}^{M}S^l.
\end{equation}

The basic idea of the proposed WEAC method is to construct the association matrix with considering the reliability of the base clusterings. We assess the quality of each base clustering with the NCAI measure (as described in Section~\ref{sec:cae}) and assign a weight to each base clustering with regard to its estimated quality.

\begin{mydef}
\label{def:wca}
The weighted co-association matrix $\tilde{A}$ is a $n\times n$ matrix which is computed as follows:
\begin{equation}
\label{eq:wca}
\tilde{A}=\sum_{l=1}^{M}w_lS^l,
\end{equation}
where
\begin{equation}
w_l=\frac{I_{\mathit{NCAI}}(P^l)}{\sum_{i=1}^M I_{\mathit{NCAI}}(P^i)}
\end{equation}
is the weight of the base clustering $P^l$.
\end{mydef}

According to Definitions~\ref{def:similarity_matrix} and \ref{def:wca}, for $i=1,\dots,n$ and $j=1,\dots,n$, it holds that $\tilde{A}_{ij}\in [0,1]$. Thus the labeling information of multiple base clusterings is mapped into a new similarity measure by utilizing pair-wise co-occurrence cues and reliability assessment of each member. With the weighted co-association matrix constructed, we further perform the agglomerative clustering methods \cite{jain10_survey} to achieve the final consensus clustering.

For clarity, the WEAC method is summarized in Algorithm 1.
\begin{figure}[!htb]
\textbf{Algorithm 1 (Weighted Evidence Accumulation Clustering)}\\
\small{ {\bfseries Input:} $\mathcal{P}$, $k$.
\begin{algorithmic}[1]
    \STATE Initialization: \\
    Evaluate the quality of each base clustering in $\mathcal{P}$ with NCAI according to Eq.~(\ref{eq:ncai}) and (\ref{eq:i_ncai}).
    \FOR {$l=1,2,\dots,M$}
        \STATE Construct the similarity matrix $S^l$ for $P^l$ according to Eq.~(\ref{eq:similarity_matrix}).
    \ENDFOR
    \STATE Build the weighted co-association matrix $\tilde{A}$ according to Eq.~(\ref{eq:wca}).
    \STATE Use the agglomerative methods to obtain the consensus clustering with $k$ clusters.
\end{algorithmic}
{\bfseries Output:} the consensus clustering $P^*$.}
\end{figure}

\subsection{Graph Partitioning with Multi-Granularity Link Analysis (GP-MGLA)}
\label{sec:gp_mgla}

There are three levels of granularity in the clustering ensemble, namely, the data instances, the clusters, and the base clusterings. The existing methods mainly focus on the level of data instances and that of clusters and lack the ability to treat the three levels of granularity as a whole system. In this section, we proposed a graph based clustering ensemble method termed graph partitioning with multi-granularity link analysis (GP-MGLA). In the proposed GP-MGLA method, we formulate the three levels of granularity in the clustering ensemble into a bipartite graph model, which will be described in the following.

Compared to the previous clustering ensemble methods based on graph partitioning \cite{strehl02, fern04_bipartite}, the GP-MGLA method is distinguished mainly in two aspects. Firstly, the GP-MGLA method utilizes the crowd agreement estimation mechanism (see Section~\ref{sec:cae}) for exploiting the relationship among base clusterings and evaluating the quality of the base clusterings in an unsupervised manner. Secondly, the links between clusters are integrated into the graph model via the SACT similarity measure.

In our bipartite graph model, both data instances and clusters are treated as graph nodes. There are two types of links in the graph, that is, the links between instances and the cluster containing them and the links between clusters that have common neighbors. To implement the bipartite structure, each cluster is used twice, i.e., for each cluster, there are two different nodes representing it in the bipartite graph.

Formally, we construct the bipartite graph as follows:
\begin{equation}
\label{eq:b_graph}
G=(U,V,L),
\end{equation}
where $U=\mathcal{X}\cup \mathcal{C}$ is the set of nodes including all instances and clusters, $V=\mathcal{C}$ is the set of nodes including all clusters, and $L$ is the set of graph links. The graph $G$ is an undirected graph. There are no links between the nodes in $U$ or between the nodes in $V$. All links are constructed between the nodes in $U$ and those in $V$.

Let $u_i\in U$ and $v_j\in V$ be two nodes in the graph $G$. If $u_i$ is a data instance and $v_j$ is the cluster containing $u_i$, then a link exists between $u_i$ and $v_j$ and the link between them is weighted with regard to the quality of the base clustering that $v_j$ belongs to. If both $u_i$ and $v_j$ are clusters, then the link between them is constructed via the SACT measure (see Section~\ref{sec:sact}). Formally, the weight of the link between the nodes $u_i$ and $v_j$ is defined as follows:

\begin{equation}
\label{eq:link_w}
w_{ij}=\begin{cases}
\alpha \cdot I_{\mathit{NCAI}}(P(v_j)), & \text{if} ~u_i\in\mathcal{X},v_j\in\mathcal{C}, u_i\in v_j,\\
\mathit{SIM}_{\mathit{SACT}}(u_i,v_j), & \text{if}~u_i\in\mathcal{C},v_j\in\mathcal{C},\\
0, & \text{otherwise.}
\end{cases}
\end{equation}

In the graph $G$, the instances and the clusters are used as nodes and the relationship among them is incorporated into the graph links. Also the information among the base clusterings is exploited to provide a reliability measure for the graph links via the crowd agreement estimation. With regard to the bipartite structure of the graph $G$, the  Tcut algorithm \cite{CVPR12_Li} can be utilized for partitioning the graph into a specific number of disjoint sets of nodes. The data instances in each of these disjoint sets are treated as a cluster and thus the final consensus clustering is obtained. Theoretically, there is a possibility that some of these disjoint sets consist of only clusters and no instances, which would lead to a less number of clusters than specified. However, we have never come across this situation in our experiments, probably due to that the joint force of the links between the instances and clusters containing them is strong enough to hold at least part of them together. For clarity, we summarize the GP-MGLA method in Algorithm 2.

\begin{figure}[!htb]
\textbf{Algorithm 2 (Graph Partitioning with Multi-Granularity Link Analysis)}\\
\small{ {\bfseries Input:} $\mathcal{P}$, $k$.
\begin{algorithmic}[1]
    \STATE Initialization: \\
    Evaluate the quality of each base clustering in $\mathcal{P}$ with NCAI according to Eq.~(\ref{eq:ncai}) and (\ref{eq:i_ncai}).\\
    Compute the SACT similarity between clusters according to Eq.~(\ref{eq:sact_2}).
    \STATE Build the bipartite graph $G=(U,V,L)$ with $U=\mathcal{X}\cup \mathcal{C}$, $V=\mathcal{C}$, and $L$ constructed as Eq.~(\ref{eq:link_w}).
    \STATE Partition the graph $G$ with the Tcut algorithm into $k$ disjoint sets of nodes.
    \STATE Treat the data instances in each set as a cluster and thus obtain the consensus clustering.
\end{algorithmic}
{\bfseries Output:} the consensus clustering $P^*$.}
\end{figure}

\section{Experiments}
\label{sec:experiment}
In this section, we conduct experiments on eight real-world datasets and compare the proposed approaches against several baseline clustering ensemble approaches. The datasets and evaluation criterion are described in Section~\ref{sec:dataset}. The setting of parameters is discussed in Sections~\ref{sec:para_analysis}. The construction of base clusterings is introduced in Section~\ref{sec:bc_pool}. Then we evaluate the performance of the proposed methods compared to the baseline methods in Section~\ref{sec:comparison}. The analysis of computational complexity is presented in Sections~\ref{sec:time_complexity}.

The experiments in this paper are conducted in Matlab 7.14.0.739 (R2012a) 64-bit on a workstation (Windows Server 2008 R2 64-bit, 8 Intel 2.40GHz processors, 96GB of RAM).

\subsection{Datasets and Evaluation Criterion}
\label{sec:dataset}

\begin{table}[!t]
\centering
\caption{Description of the benchmark datasets}
\label{table:datasets}
\begin{center}\vskip -0.1 in
\begin{tabular}{p{4cm}<{\centering}p{3cm}<{\centering}p{3cm}<{\centering}p{2.5cm}<{\centering}}
\toprule
Dataset         &\#Instance     &\#Attribute      &\#Class\\
\midrule
\emph{Breast Cancer}        &683      &9    &2\\
\emph{Image Segmentation}   &2,310  &19     &7\\
\emph{Iris}                 &150    &4      &3\\
\emph{Seeds}                &210    &7      &3\\
\emph{Yeast}                &1,484  &8      &10\\
\emph{Wine}                 &178    &13     &3\\
\emph{Pen Digits}           &10,992 &16     &10\\
\emph{Letters}              &20,000 &16     &26\\
\bottomrule
\end{tabular}
\end{center}
\end{table}

In our experiments, eight real-world datasets from the UCI machine learning repository \cite{Bache+Lichman:2013} are used, namely, \emph{Breast Cancer}, \emph{Image Segmentation}, \emph{Iris}, \emph{Seeds}, \emph{Yeast}, \emph{Wine}, \emph{Pen Digits}, and \emph{Letters}. The details of the benchmark datasets are given in \tablename~\ref{table:datasets}.

To evaluate the quality of the consensus clustering, we utilize the normalized mutual information (NMI) \cite{strehl02} which provides an indication of the shared information between two clusterings. Let $P^*$ be the test clustering and $P^G$ the ground-truth clustering. The NMI score of $P^*$ w.r.t. $P^G$ is computed as follows:
\begin{equation}
\label{eq:nmi}
NMI(P^*, P^G)=\frac{\sum_{i=1}^{n^*}\sum_{j=1}^{n^G}n_{ij}\log\frac{n_{ij}n}{n_i^*n_j^G}}{\sqrt{\sum_{i=1}^{n^*}n_i^*\log\frac{n_i^*}{n}\sum_{j=1}^{n^G}n_j^G\log\frac{n_j^G}{n}}},
\end{equation}
where $n^*$ is the number of clusters in $P^*$, $n^G$ is the number of clusters in $P^G$, $n_i^*$ is the number of instances in the $i$-th cluster of $P^*$, $n_j^G$ is the number of instances in the $j$-th cluster of $P^G$, and $n_{ij}$ is the number of common instances shared by cluster $i$ in $P^*$ and cluster $j$ in $P^G$.

\subsection{Choices of Parameters}
\label{sec:para_analysis}

There is one parameter $\beta$ in the WEAC method and two parameters $\alpha$ and $\beta$ in the GP-MGLA method. The parameter $\alpha$ is a scale factor for the link weights between instances and clusters. The parameter $\beta$ adjusts the influence of NCAI for both WEAC and GP-MGLA, where a bigger $\beta$ signals a greater influence of NCAI. We evaluate the performance of the proposed WEAC and GP-MGLA methods with varying parameters on the benchmark datasets. As can be seen in \tablename~\ref{table:para_weac} and \ref{table:para_gp_mgla}, the proposed methods are very stable w.r.t. the varying parameters. Empirically, it is suggested that $\alpha$ be set in the interval of $(0.1,1)$ and $\beta$ be set in the interval of $(1,4)$ for the proposed two methods. In the following, the parameters are set that $\alpha=0.5$ and $\beta=2$ for all the experiments on all the benchmark datasets.

\begin{table}[!t]
\centering
\caption{The performance of WEAC with varying parameters in terms of NMI}
\label{table:para_weac}
\begin{tabular}{|m{4cm}<{\centering}|m{1.10cm}<{\centering}m{1.10cm}<{\centering}m{1.10cm}<{\centering}m{1.10cm}<{\centering}m{1.10cm}<{\centering}|}
\hline
\multirow{2}{*}{Dataset}             &\multicolumn{5}{c|}{$\beta$}\\
\cline{2-6}
             &0          &1          &2          &4          &8          \\
\hline
\emph{Breast Cancer}      &0.647           &0.673      &0.674      &0.687      &0.685      \\
\emph{Iris}             &0.734        &0.743           &0.778      &0.748      &0.750      \\
\emph{Image Segmentation} &0.639          &0.641      &0.648      &0.647      &0.657      \\
\emph{Seeds}          &0.591    &0.623      &0.634           &0.624      &0.626      \\
\emph{Yeast}          &0.230   &0.234      &0.232           &0.241      &0.239\\
\emph{Wine}           &0.753   &0.772      &0.781           &0.781      &0.757\\
\emph{Pen Digits}           &0.742	&0.753	&0.770	&0.777	&0.796\\
\emph{Letters}              &0.434	&0.443	&0.444	&0.451	&0.454\\
\hline
\end{tabular}
\end{table}

\begin{table}[!t]
\centering
\caption{The performance of GP-MGLA with varying parameters in terms of NMI}
\label{table:para_gp_mgla}
\begin{tabular}{|m{3.7cm}<{\centering}|m{0.93cm}<{\centering}m{0.93cm}<{\centering}m{0.93cm}<{\centering}m{0.93cm}<{\centering}|m{0.93cm}<{\centering}m{0.93cm}<{\centering}m{0.93cm}<{\centering}|}
\hline
$\alpha$            &\multicolumn{4}{c|}{0.5}       &0.01                 &0.1                        &1\\
\hline
$\beta$             &0                    &2          &4          &8          &\multicolumn{3}{c|}{2}\\
\hline
\emph{Breast Cancer}      &0.677           &0.719      &0.725      &0.729      &0.702      &0.712     &0.713\\
\emph{Iris}             &0.739        &0.742           &0.742      &0.751      &0.743      &0.748      &0.742\\
\emph{Image Segmentation} &0.635          &0.650      &0.648      &0.649      &0.639      &0.642      &0.651\\
\emph{Seeds}          &0.593    &0.620      &0.621           &0.609      &0.611      &0.614      &0.623\\
\emph{Yeast}          &0.239   &0.251      &0.252           &0.250      &0.243      &0.246      &0.249\\
\emph{Wine}           &0.781   &0.798      &0.792           &0.783      &0.788      &0.786      &0.794\\
\emph{Pen Digits}           &0.779	&0.796	&0.803	&0.800    &0.788	&0.792	&0.798\\
\emph{Letters}              &0.448	&0.456	&0.456	&0.461    &0.449	&0.454	&0.456\\
\hline
\end{tabular}
\end{table}

\subsection{Generation of Base Clustering Ensemble}
\label{sec:bc_pool}
The proposed approaches make no specific assumption about the generation of the ensemble of base clusterings. To evaluate the effectiveness and robustness of the proposed methods over various combinations of base clusterings, we construct a pool of a large number of different base clusterings. Then we run the proposed methods and the baseline methods with the base clusterings randomly chosen from the pool repeatedly.

Four clustering algorithms are used to construct the base clustering pool, namely, $k$-means, rival penalized competitive learning (RPCL) \cite{xu93_rpcl}, hierarchical mode association clustering (HMAC) \cite{li07_hmac}, and incremental support vector clustering with outlier detection (OD-ISVC) \cite{huang2012incremental}. To obtain a pool of various base clusterings, we apply the aforementioned clustering algorithms repeatedly with random parameters and initializations on each dataset. The number of clusters for the $k$-means and RPCL methods are randomly chosen in the interval of $[2,2\sqrt{n}]$, where $n$ is the number of instances in the dataset. The HMAC method is a hierarchical clustering method. We choose the hierarchy of clustering randomly for the HMAC method, where each hierarchy corresponds to a clustering with a certain number of clusters. For the OD-ISVC method, the base clusterings are generated with randomly chosen kernel width parameter $q$ and trade-off parameter $C$. In this paper, we apply each of the clustering algorithms for $100$ times and thus a pool of $400$ different base clusterings is constructed for each dataset.

\subsection{Performance Comparison and Analysis}
\label{sec:comparison}
With the base clustering pool constructed (see Section~\ref{sec:bc_pool}), the proposed approaches and the baseline approaches are applied to the ensemble of base clusterings which are randomly chosen from the pool. In our experiments, each of the clustering ensemble approaches has no knowledge about how the chosen base clusterings are generated, i.e., by which algorithm and with what parameters they are generated. For each run, an ensemble of $M$ base clusterings is randomly constructed and different clustering ensemble approaches are applied to the ensemble. The ensemble size $M=5$ is used in our work. We test the proposed approaches against the baseline approaches by evaluating their performance over a large number of runs, which aims to rule out the factor of ``\emph{getting lucky sometimes}'' and provide a fair comparison for their effectiveness and robustness over different combinations of base clusterings.

\subsubsection{Comparison with Base Clusterings}

In this paper we propose two novel consensus functions, namely, the GP-MGLA method and the WEAC method. GP-MGLA is a graph partitioning based method, whereas WEAC is a pair-wise similarity based method. With the weighted co-association matrix computed by WEAC, we further perform three agglomerative methods, namely, average-link (AL), complete-link (CL), and single-link (SL) to obtain the final consensus clustering, which leads to three sub-methods denoted as WEAC-AL, WEAC-CL, and WEAC-SL respectively.

\begin{figure*}[!t]
\hskip 0.2in
\begin{center}
{
{\includegraphics[width=0.999\columnwidth]{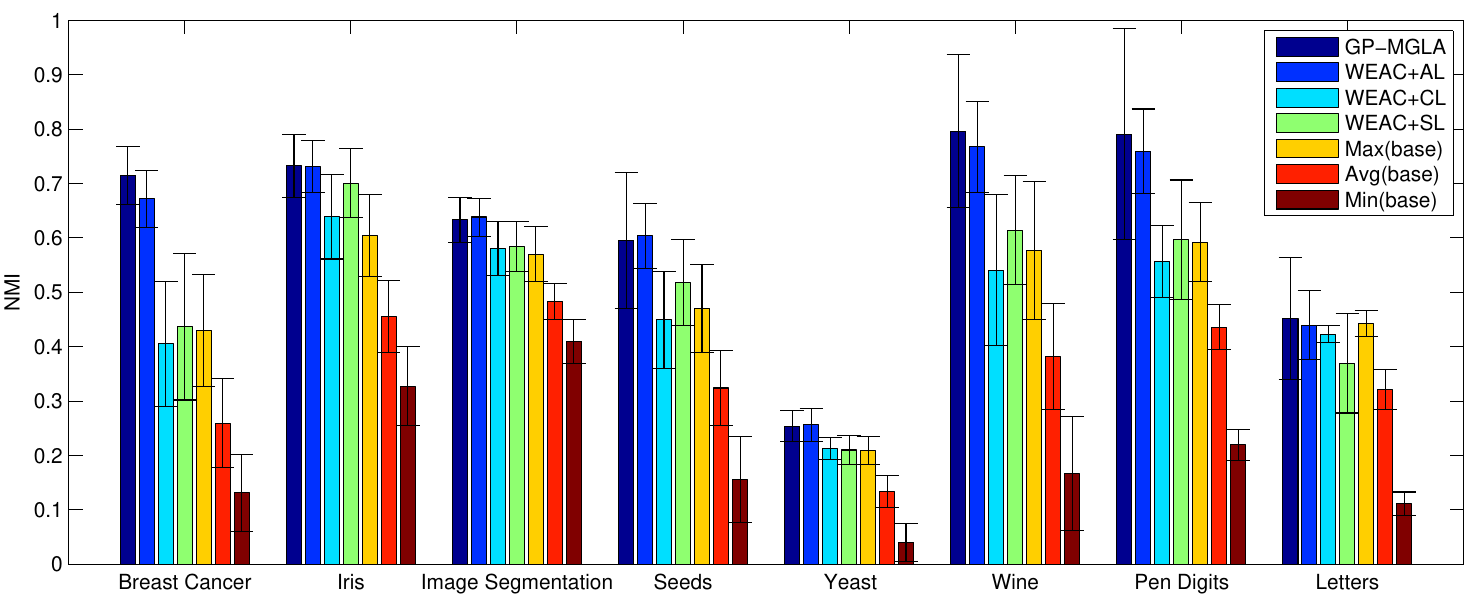}}}
\caption{Average performance in terms of NMI over 100 runs by WEAC and GP-MGLA compared to the base clusterings. }
\label{fig:comp_base}
\end{center}
\vskip -0.2in
\end{figure*}

For each run, an ensemble is generated by randomly drawing $M$ base clusterings from the pool. We apply the proposed clustering ensemble methods on different ensembles for each dataset repeatedly. The average performance over 100 runs of our methods compared to the base clusterings is shown in Fig.~\ref{fig:comp_base}, in which Max(base) denotes the average NMI score of the best base clustering over all ensembles, Min(base) denotes the average NMI score of the worst base clustering over all ensembles, and Avg(base) denotes the average NMI score of all base clusterings over all ensembles. As shown in Fig.~\ref{fig:comp_base}, the proposed methods are able to produce better and more robust consensus clusterings than the base clusterings. Specially, GP-MGLA and WEAC-AL significantly outperform the base clusterings on the \emph{Breast Cancer}, \emph{Seeds}, \emph{Wine}, and \emph{Pen Digits} datasets.

We further compare the consensus clusterings by our methods against each of the base clusterings and calculate the winning percentage. For each run, an ensemble of $M$ base clusterings are selected. Then there will be totally $100\cdot M$ comparisons between the consensus clusterings and the base clusterings over 100 runs. We call it a \emph{win} if the consensus clustering has a higher NMI score than a base clustering and call it a \emph{loss} if the consensus clustering has a lower NMI score than a base clustering. Ties count as $1/2$ win and $1/2$ loss. The winning percentage is defined as the number of wins divided by the total number of comparisons. As shown in \tablename~\ref{table:win_percent_bestk}, the GP-MGLA method and the WEAC method (associated with AL) outperform most of the base clustering w.r.t. the best number of clusters on the benchmark datasets. We also compare the consensus clusterings against the base clusterings w.r.t. the same number of clusters, which means for each comparison the number of clusters of the consensus clustering are set to the same number as the base clustering. As shown in \tablename~\ref{table:win_percent_samek}, GP-MGLA and WEAC outperform about two thirds of the base clusterings w.r.t. the same number of clusters on the benchmark datasets.

\begin{table}[!t]
\centering
\caption{Winning percentage of the consensus clustering against base clusterings w.r.t. the best number of clusters over 100 runs.}
\label{table:win_percent_bestk}
\begin{center}\vskip -0.1 in
\begin{tabular}{|c|c|c|c|c|}
\hline
Dataset         &\emph{Breast Cancer}     &\emph{Image Segmentation}      &\emph{Iris}      &\emph{Seeds}\\
\hline
GP-MGLA         &99.6\%&97.2\%&98.4\%&99.2\%\\
\hline
WEAC            &99.0\%&99.2\%&98.4\%&99.2\%\\
\hline
\hline
Dataset         &\emph{Yeast}     &\emph{Wine}      &\emph{Pen Digits}      &\emph{Letters}\\
\hline
GP-MGLA         &100\%&98.0\%&100\%&99.0\%\\
\hline
WEAC            &100\%&96.8\%&100\%&70.6\%\\
\hline
\end{tabular}
\end{center}
\end{table}
\begin{table}[!t]
\centering
\caption{Winning percentage of the consensus clustering against base clusterings w.r.t. the same number of clusters over 100 runs.}
\label{table:win_percent_samek}
\begin{center}\vskip -0.1 in
\begin{tabular}{|c|c|c|c|c|}
\hline
Dataset         &\emph{Breast Cancer}     &\emph{Image Segmentation}      &\emph{Iris}      &\emph{Seeds}\\
\hline
GP-MGLA         &59.7\%&72.5\%&70.7\%&65.8\%\\
\hline
WEAC            &67.4\%&76.0\%&70.0\%&71.1\%\\
\hline
\hline
Dataset         &\emph{Yeast}     &\emph{Wine}      &\emph{Pen Digits}      &\emph{Letters}\\
\hline
GP-MGLA         &64.5\%&68.1\%&95.7\%&73.0\%\\
\hline
WEAC            &68.1\%&73.8\%&97.3\%&66.6\%\\
\hline
\end{tabular}
\end{center}
\end{table}

\subsubsection{Comparison with Other Clustering Ensemble Methods}

\begin{table}[!thb]\footnotesize
\centering 
\caption{Average performance in terms of NMI over 100 runs by different clustering ensemble methods (The two highest scores in each column are highlighted in bold.)}
\label{table:compare_ce}
\begin{tabular}{|m{2.5cm}<{\centering}|m{1.25cm}<{\centering}m{1.25cm}<{\centering}|m{1.25cm}<{\centering}m{1.25cm}<{\centering}|m{1.6cm}<{\centering}m{1.6cm}<{\centering}|}
\hline
\multirow{2}{*}{Method}               &\multicolumn{2}{c|}{\emph{Breast Cancer}}&\multicolumn{2}{c|}{\emph{Iris}}   &\multicolumn{2}{c|}{\emph{Image Segmentation}}\\
\cline{2-7}
&Best-$k$&True-$k$    &Best-$k$&True-$k$    &Best-$k$&True-$k$\\
\hline
GP-MGLA    &\textbf{0.715}&\textbf{0.618}    &\textbf{0.733}&\textbf{0.695}     &0.633&\textbf{0.549}\\
\hline
WEAC+AL                      &0.672&0.596    &\textbf{0.731}&\textbf{0.673}     &\textbf{0.638}&\textbf{0.533}\\
WEAC+CL                      &0.405&0.073    &0.639&0.653                       &0.580&0.317\\
WEAC+SL                      &0.437&0.030    &0.701&0.493                       &0.584&0.420\\
\hline
HBGF       &\textbf{0.695}&\textbf{0.648}    &0.707&0.640                       &0.631&0.491\\
\hline
WCC                         &0.621&0.459    &0.694&0.539                       &0.621&0.527\\
\hline
EAC+AL                      &0.652&0.512    &0.725&0.667                        &0.637&0.503\\
EAC+CL                      &0.421&0.058    &0.637&0.497                        &0.582&0.323\\
EAC+SL                      &0.377&0.010    &0.680&0.632                        &0.535&0.413\\
\hline
ECMC+AL                      &0.436&0.399    &0.272&0.140                        &0.100&0.081\\
ECMC+CL                      &0.390&0.358    &0.306&0.181                        &0.126&0.102\\
ECMC+SL                      &0.381&0.259    &0.403&0.272                        &0.060&0.026\\
\hline
SRS+AL                      &0.650&0.519    &0.726&0.676                        &\textbf{0.642}&0.513\\
SRS+CL                      &0.632&0.489    &0.708&0.648                        &0.624&0.530\\
SRS+SL                      &0.544&0.029    &0.706&0.661                        &0.619&0.411\\
\hline
WCT+AL                      &0.668&0.075    &0.724&\textbf{0.673}               &0.632&0.494\\
WCT+CL                      &0.621&0.110    &0.698&0.644                        &0.615&0.492\\
WCT+SL                      &0.546&0.124    &0.705&0.650                        &0.580&0.416\\
\hline
\hline
\multirow{2}{*}{Method}               &\multicolumn{2}{c|}{\emph{Seeds}}&\multicolumn{2}{c|}{\emph{Yeast}}   &\multicolumn{2}{c|}{\emph{Wine}}\\
\cline{2-7}
&Best-$k$&True-$k$    &Best-$k$&True-$k$    &Best-$k$&True-$k$\\
\hline
GP-MGLA   &\textbf{0.595}&\textbf{0.514}    &0.254&0.167  &\textbf{0.797}&\textbf{0.717}\\
\hline
WEAC+AL   &\textbf{0.604}&\textbf{0.517}    &\textbf{0.256}&0.147                    &0.767&\textbf{0.664}\\
WEAC+CL                     &0.449&0.197    &0.212&0.093                    &0.540&0.177\\
WEAC+SL                     &0.517&0.317    &0.210&0.046                    &0.614&0.235\\
\hline
HBGF                        &0.587&0.493    &\textbf{0.256}&\textbf{0.181}           &\textbf{0.781}&0.647\\
\hline
WCC                        &0.567&0.439    &0.245&\textbf{0.208}          &0.701&0.581\\
\hline
EAC+AL                      &0.582&0.399    &\textbf{0.256}&0.109                    &0.733&0.444\\
EAC+CL                      &0.467&0.206    &0.218&0.065                    &0.547&0.168\\
EAC+SL                      &0.502&0.244    &0.173&0.034                    &0.580&0.104\\
\hline
ECMC+AL                      &0.233&0.126    &0.073&0.021                        &0.270&0.154\\
ECMC+CL                      &0.238&0.146    &0.074&0.022                        &0.266&0.159\\
ECMC+SL                      &0.216&0.064    &0.073&0.017                        &0.253&0.078\\
\hline
SRS+AL                      &0.584&0.438    &\textbf{0.256}&0.122                    &0.733&0.254\\
SRS+CL                      &0.547&0.356    &0.229&0.116                    &0.658&0.407\\
SRS+SL                      &0.560&0.344    &0.204&0.034                    &0.668&0.001\\
\hline
WCT+AL                      &0.573&0.403    &\textbf{0.268}&0.120           &0.730&0.478\\
WCT+CL                      &0.555&0.326    &0.230&0.095                    &0.672&0.396\\
WCT+SL                      &0.528&0.289    &0.221&0.035                    &0.639&0.184\\
\hline
\end{tabular}
\end{table}

\begin{table}[!thb]\footnotesize
\centering 
\caption{Average performance in terms of NMI over 100 runs by different clustering ensemble methods (The two highest scores in each column are highlighted in bold.)}
\label{table:compare_ce2}
\begin{tabular}{|m{2.5cm}<{\centering}|m{1.25cm}<{\centering}m{1.25cm}<{\centering}|m{1.25cm}<{\centering}m{1.25cm}<{\centering}|}
\hline
\multirow{2}{*}{Method}               &\multicolumn{2}{c|}{\emph{Pen Digits}}&\multicolumn{2}{c|}{\emph{Letters}}\\
\cline{2-5}
&Best-$k$&True-$k$    &Best-$k$&True-$k$\\
\hline
GP-MGLA    &\textbf{0.791}&\textbf{0.725}    &\textbf{0.452}&\textbf{0.378}\\
\hline
WEAC+AL             &0.760&\textbf{0.667}    &0.440&0.345\\
WEAC+CL                      &0.557&0.243    &0.423&0.160\\
WEAC+SL                      &0.597&0.167    &0.370&0.092\\
\hline
HBGF                &\textbf{0.781}&0.663    &\textbf{0.445}&\textbf{0.364}\\
\hline
EAC+AL                      &0.745&0.588    &0.425&0.290\\
EAC+CL                      &0.584&0.244    &0.322&0.139\\
EAC+SL                      &0.445&0.085    &0.102&0.061\\
\hline
SRS+AL                      &0.750&0.611    &0.424&0.300\\
SRS+CL                      &0.684&0.485    &0.391&0.280\\
SRS+SL                      &0.706&0.127    &0.070&0.032\\
\hline
WCT+AL                      &0.765&0.584    &0.442&0.333\\
WCT+CL                      &0.661&0.392    &0.414&0.278\\
WCT+SL                      &0.718&0.136    &0.121&0.065\\
\hline
\end{tabular}
\end{table}

We compare the proposed WEAC and GP-MGLA methods against six different clustering ensemble methods, namely, the hybrid bipartite graph formulation (HBGF) \cite{fern04_bipartite}, the weighted consensus clustering (WCC) \cite{Li_WCC08}, the evidence accumulation clustering (EAC) \cite{Fred05_EAC}, the ensemble clustering by matrix completion (ECMC) \cite{yi_icdm12}, the SimRank similarity based method (SRS) \cite{iamon08_icds}, and the weighted connected-triple method (WCT) \cite{iam_on11_linkbased}. Since the ECMC method and the WCC method is very time-consuming (see Fig.~\ref{fig:time}), it is almost infeasible to run ECMC and WCC for 100 times on the large datasets as the \emph{Pen Digits} and \emph{Letters} datasets, which contain $10,992$ and $20,000$ instances respectively. Therefore, the ECMC and WCC methods are performed on the benchmark datasets except the \emph{Pen Digits} and \emph{Letters} datasets.  And the other baseline methods are performed on all the benchmark datasets.

The EAC, ECMC, SRS, and WCT methods are four pair-wise similarity based methods, each leading to three sub-methods by utilizing three different agglomerative clustering methods, namely, AL, CL, and SL. Then we have 14 baseline methods, that is, HBGF, WCC, EAC-AL, EAC-CL, EAC-SL, ECMC-AL, ECMC-CL, ECMC-SL, SRS-AL, SRS-CL, SRS-SL, WCT-AL, WCT-CL, and WCT-SL. The average performance over 100 runs of the proposed methods and the 14 baseline methods for each dataset is summarized in \tablename~\ref{table:compare_ce} and \ref{table:compare_ce2}. For each test method, the number of clusters $k$ for the consensus clustering is set to two values respectively, that is, best-$k$ and true-$k$. The best-$k$ is the number of clusters that leads to the optimal performance for a method on the dataset. The true-$k$ is the number of true classes in the dataset. As shown in \tablename~\ref{table:compare_ce} and \ref{table:compare_ce2}, the performance of the WEAC-AL method is better and more stable than the other pair-wise similarity based methods. The WEAC-AL method achieves the best NMI scores for the \emph{Seeds} dataset and nearly best NMI scores for the \emph{Iris}, \emph{Image Segmentation}, \emph{Yeast}, and \emph{Wine} datasets. Among the test methods, the GP-MGLA method produces overall the best and most stable clustering results on the benchmark datasets.

\subsubsection{Dealing with Ill Clusterings}

\begin{figure*}[!t]
\hskip 0.2in
\begin{center}
{\subfigure[\emph{Breast Cancer}]
{\includegraphics[width=0.32\columnwidth]{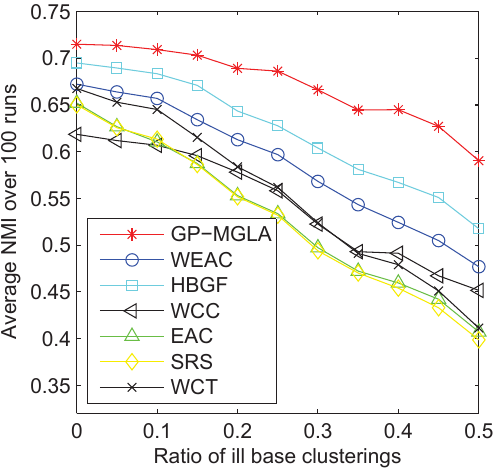}}}
{\subfigure[\emph{Iris}]
{\includegraphics[width=0.32\columnwidth]{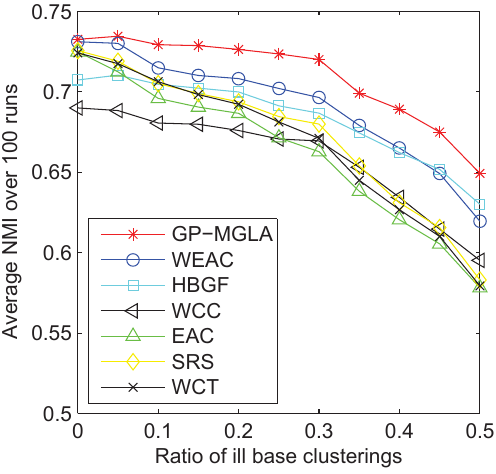}}}
{\subfigure[\emph{Image Segmentation}]
{\includegraphics[width=0.32\columnwidth]{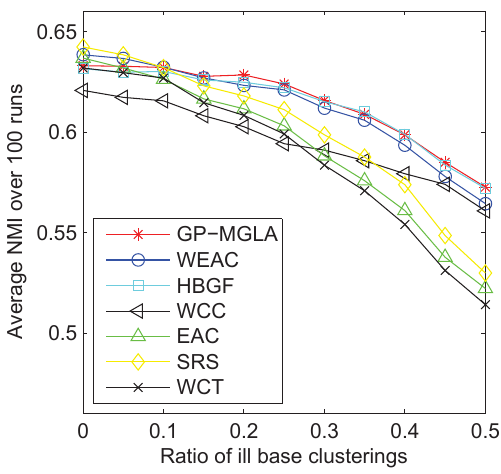}}}
{\subfigure[\emph{Seeds}]
{\includegraphics[width=0.32\columnwidth]{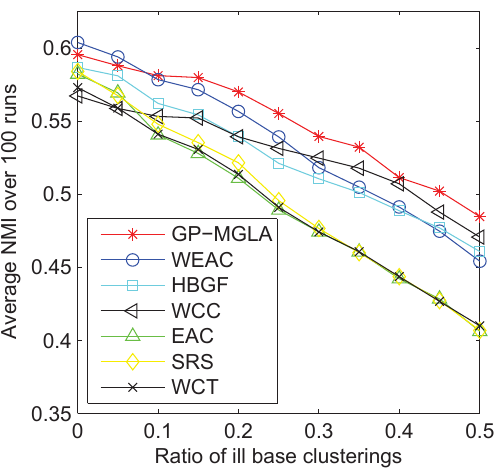}}}
{\subfigure[\emph{Yeast}]
{\includegraphics[width=0.32\columnwidth]{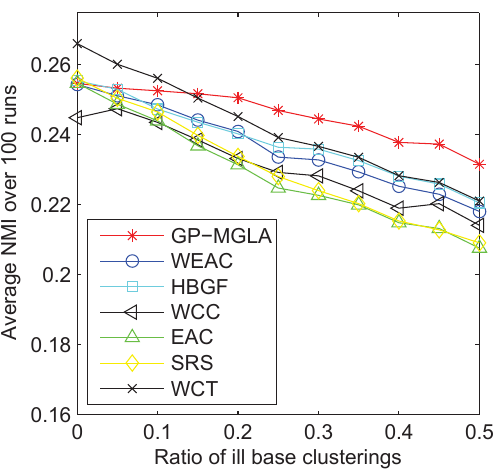}}}
{\subfigure[\emph{Wine}]
{\includegraphics[width=0.32\columnwidth]{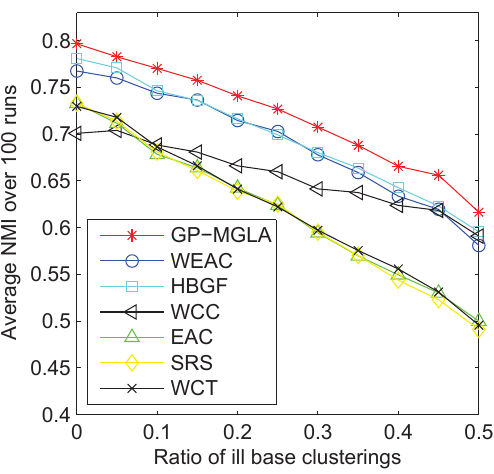}}}
{\subfigure[\emph{Pen Digits}]
{\includegraphics[width=0.32\columnwidth]{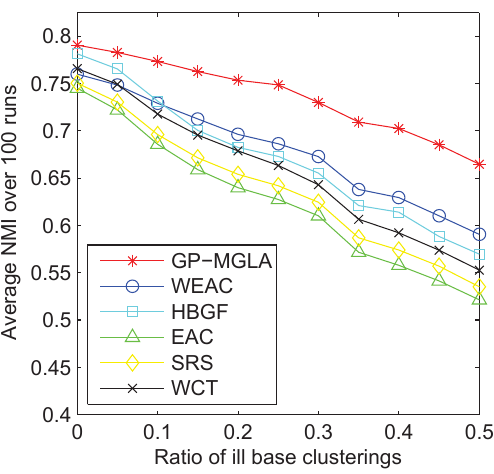}}}
{\subfigure[\emph{Letters}]
{\includegraphics[width=0.32\columnwidth]{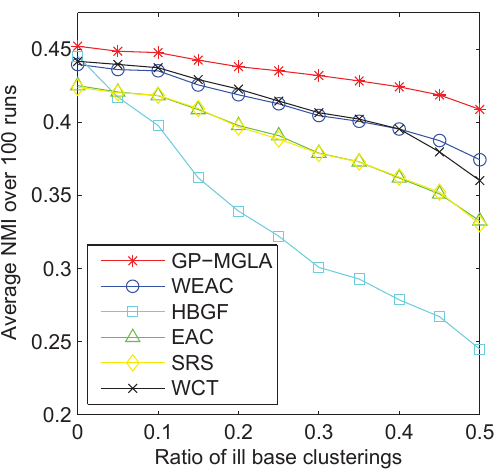}}}
\caption{The performance by the proposed methods and the baseline methods over different ratio of ill base clusterings.}
\label{fig:comp_ill}
\end{center}
\vskip -0.2in
\end{figure*}

In order to evaluate the robustness of our methods to ill clusterings, we add a certain ratio of heavily imbalanced clusterings into the base clustering pool. For example, adding $20\%$ of ill clusterings into the pool means replacing $20\%$ of base clusterings in the pool with heavily imbalanced clusterings. To produce the heavily imbalanced clusterings, we firstly partition the dataset into $k$ clusters via $k$-means where $k$ is randomly chosen in the interval of $[\sqrt{n},2\sqrt{n}]$. Then we merge a proportion $\rho$ of clusters into one, i.e., $\rho\cdot k$ randomly chosen clusters will be merged into one cluster in the clustering. In our experiments, the values of $\rho$ are randomly selected in the interval of $(0.7,0.99)$, which lead to heavily imbalanced clusterings. Different ratio of ill base clusterings are added to the pool and then we conduct experiments on the ensemble of randomly chosen base clusterings from the pool.

For each ratio of ill base clusterings, we run each of the clustering ensemble methods for 100 times and the performance is summarized in Fig.~\ref{fig:comp_ill}. The average-link is used for each of the pair-wise similarity based methods, namely, WEAC, EAC, ECMC, SRS, and WCT. As can be seen in Fig.~\ref{fig:comp_ill}, the proposed WEAC method yields much better performance than the EAC method on the benchmark datasets. On the whole, the proposed GP-MGLA method yields much better and more robust performance than the other clustering ensemble methods with different ratio of ill base clusterings added.

\subsection{Computational Complexity}
\label{sec:time_complexity}

The computation of the NMI measure between two partitions takes $O(n^2)$ time, where $n$ is the number of instances in the dataset. The computation of the NCAI measure takes $O(M^2n^2)$ time, where $M$ is the number of base clusterings in the ensemble. The computation of the SACT similarity is $O(M^2n^2+l{n_c}^2+nn_c)$, where $n_c$ is the number of clusters in the ensemble and $l$ is the average number of neighbors connecting to a cluster. As the conventional EAC method \cite{Fred05_EAC} is $O(Mn^2)$, the time complexity of the proposed WEAC method (associated with average-link) is $O(M^2n^2)$. The Tcut algorithm for bipartite graph partitioning is $O(kdn+k{n_c}^2)$, where $k$ is the number of clusters in the final consensus clustering and $d$ is the average number of links connecting to a node in the graph. Then we have the time complexity of the proposed GP-MGLA method as $O(M^2n^2+(l+k){n_c}^2+nn_c+kdn)$.

\begin{figure*}[!tb]
\label{fig:time_complexity}
\hskip 0.2in
\begin{center}
{\subfigure[]
{\includegraphics[width=0.4\columnwidth]{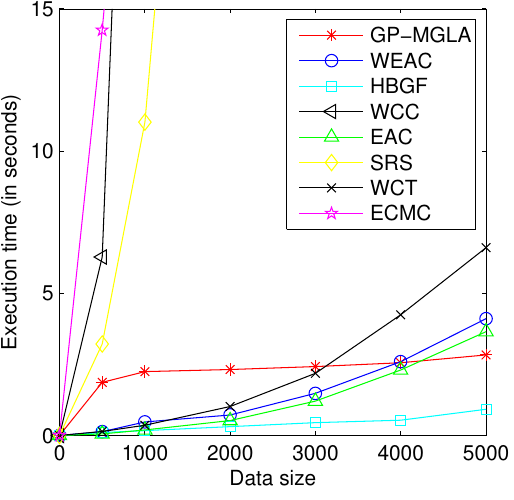}}}
{\subfigure[]
{\includegraphics[width=0.4\columnwidth]{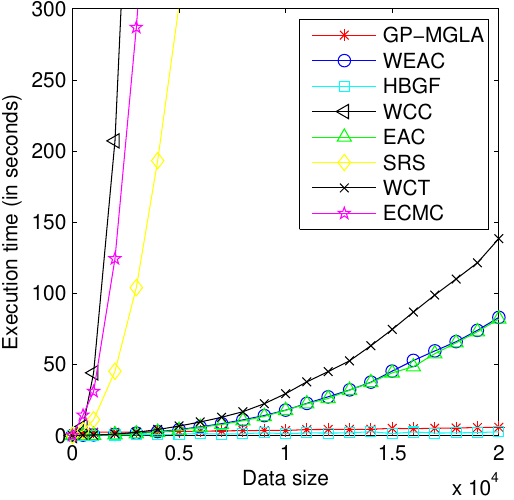}}}
\caption{Execution time of different clustering ensemble approaches as the data size varies (a) from 0 to 5,000 and (b) from 0 to 20,000. }
\label{fig:time}
\end{center}
\vskip -0.2in
\end{figure*}

The proposed methods and the baseline methods are applied to the \emph{Letters} dataset to test the execution time w.r.t. varying data sizes. The time performance of these test methods with varying data sizes is illustrated in Fig.~\ref{fig:time}. To process the entire \emph{Letters} dataset with $20,000$ instances, the time costs (in seconds) of WEAC and GP-MGLA are $82.94$ and $5.64$ respectively, whereas the time costs (in seconds) of HBGF, EAC, and WCT are $2.51$, $81.91$, and $138.43$ respecitively. In the proposed methods, it takes $2.01$ seconds to compute the NCAI for the data size of $20,000$. Each of the five pair-wise similarity based methods, namely, WEAC, EAC, ECMC, SRS, and WCT, is associated with average-link. As shown in Fig.~\ref{fig:time}, the ECMC method and the WCC method are the two slowest methods. And the SRS method is the third slowest. The GP-MGLA is slower than WEAC, EAC, and WCT when the data size is below $4,000$. However, the GP-MGLA shows an advantage in execution time as the data size grows beyond $5,000$. The proposed GP-MGLA method and the HBGF method are the two fastest methods when the data size is greater than $5,000$, mainly due to their efficient graph partitioning algorithms.

\section{Conclusions}
\label{sec:conclusion}

In this paper, we address the clustering ensemble problem using crowd agreement estimation and multi-granularity link analysis. With the clustering ensemble viewed as a crowd, we assess reliability of the individuals inside it by exploiting the so-called wisdom of the crowd. The normalized crowd agreement index is proposed for evaluating the quality of base clusterings in an unsupervised manner. The source aware connected triple similarity is introduced for constructing the link between two clusters with their common neighbors and source reliability taken into consideration. To achieve the final consensus clustering, two novel consensus functions are further presented, termed weighted evidence accumulation clustering (WEAC) and graph partitioning with multi-granularity link analysis (GP-MGLA) respectively. The experiments conducted on eight real-world datasets show the effectiveness and robustness of the proposed clustering ensemble methods.

\section*{Acknowledgment}
\label{sec:acknowledgment}

The authors would like to thank the anonymous reviewers for their insightful comments and suggestions which helped enhance this paper significantly. This work was supported by National Science \& Technology Pillar Program (No. 2012BAK16B06), NSFC (61173084) and the GuangDong Program (Grant No. 2012A080104005). The work of Chang-Dong Wang was in part sponsored by CCF-Tencent Open Research Fund and Research Training Program of SMIE of Sun Yat-sen University.

\bibliographystyle{model1-num-names}
\bibliography{nc_2013_full}

\begin{thebibliography}{39}
\expandafter\ifx\csname natexlab\endcsname\relax\def\natexlab#1{#1}\fi
\providecommand{\bibinfo}[2]{#2}
\ifx\xfnm\relax \def\xfnm[#1]{\unskip,\space#1}\fi
\bibitem[{Xu et~al.(1993)Xu, Krzyzak, and Oja}]{xu93_rpcl}
\bibinfo{author}{L.~Xu}, \bibinfo{author}{A.~Krzyzak},
  \bibinfo{author}{E.~Oja},
\newblock \bibinfo{title}{Rival penalized competitive learning for clustering
  analysis, {RBF} net, and curve detection},
\newblock \bibinfo{journal}{IEEE Transactions on Neural Networks}
  \bibinfo{volume}{4 (4)} (\bibinfo{year}{1993}) \bibinfo{pages}{636--649}.
\bibitem[{Li et~al.(2007)Li, Ray, and Lindsay}]{li07_hmac}
\bibinfo{author}{J.~Li}, \bibinfo{author}{S.~Ray}, \bibinfo{author}{B.~G.
  Lindsay},
\newblock \bibinfo{title}{A nonparametric statistical approach to clustering
  via mode identification},
\newblock \bibinfo{journal}{Journal of Machine Learning Research}
  \bibinfo{volume}{8} (\bibinfo{year}{2007}) \bibinfo{pages}{1687--1723}.
\bibitem[{Zhang and Zhou(2009)}]{zhang09_ai}
\bibinfo{author}{M.-L. Zhang}, \bibinfo{author}{Z.-H. Zhou},
\newblock \bibinfo{title}{Multi-instance clustering with applications to
  multi-instance prediction},
\newblock \bibinfo{journal}{Applied Intelligence} \bibinfo{volume}{31 (1)}
  (\bibinfo{year}{2009}) \bibinfo{pages}{47--68}.
\bibitem[{Zhao et~al.(2010)Zhao, Jiao, Liu, Gao, and Gong}]{zhao10_nc}
\bibinfo{author}{F.~Zhao}, \bibinfo{author}{L.~Jiao}, \bibinfo{author}{H.~Liu},
  \bibinfo{author}{X.~Gao}, \bibinfo{author}{M.~Gong},
\newblock \bibinfo{title}{Spectral clustering with eigenvector selection based
  on entropy ranking},
\newblock \bibinfo{journal}{Neurocomputing} \bibinfo{volume}{73 (10-12)}
  (\bibinfo{year}{2010}) \bibinfo{pages}{1704--1717}.
\bibitem[{Wang and Lai(2011)}]{ebcl11}
\bibinfo{author}{C.-D. Wang}, \bibinfo{author}{J.-H. Lai},
\newblock \bibinfo{title}{Energy based competitive learning},
\newblock \bibinfo{journal}{Neurocomputing} \bibinfo{volume}{74 (12-13)}
  (\bibinfo{year}{2011}) \bibinfo{pages}{2265--2275}.
\bibitem[{Li et~al.(2011)Li, Lian, Kwok, and Lu}]{li11_cvpr}
\bibinfo{author}{M.~Li}, \bibinfo{author}{X.~C. Lian}, \bibinfo{author}{J.~T.
  Kwok}, \bibinfo{author}{B.~L. Lu},
\newblock \bibinfo{title}{Time and space efficient spectral clustering via
  column sampling},
\newblock in: \bibinfo{booktitle}{Proceedings of the IEEE Conference on
  Computer Vision and Pattern Recognition (CVPR'11), 2011}.
\bibitem[{Wang et~al.(2013{\natexlab{a}})Wang, Lai, Suen, and Zhu}]{meap13}
\bibinfo{author}{C.-D. Wang}, \bibinfo{author}{J.-H. Lai},
  \bibinfo{author}{C.~Y. Suen}, \bibinfo{author}{J.-Y. Zhu},
\newblock \bibinfo{title}{Multi-exemplar affinity propagation},
\newblock \bibinfo{journal}{IEEE Transactions on Pattern Analysis and Machine
  Intelligence} \bibinfo{volume}{35 (9)} (\bibinfo{year}{2013}{\natexlab{a}})
  \bibinfo{pages}{2223--2237}.
\bibitem[{Wang et~al.(2013{\natexlab{b}})Wang, Lai, Huang, and
  Zheng}]{svstream13}
\bibinfo{author}{C.-D. Wang}, \bibinfo{author}{J.-H. Lai},
  \bibinfo{author}{D.~Huang}, \bibinfo{author}{W.-S. Zheng},
\newblock \bibinfo{title}{{SVS}tream: A support vector based algorithm for
  clustering data streams},
\newblock \bibinfo{journal}{IEEE Transactions on Knowledge and Data
  Engineering} \bibinfo{volume}{25 (6)} (\bibinfo{year}{2013}{\natexlab{b}})
  \bibinfo{pages}{1410--1424}.
\bibitem[{Wang and Lai(2013)}]{psvdd13}
\bibinfo{author}{C.-D. Wang}, \bibinfo{author}{J.-H. Lai},
\newblock \bibinfo{title}{Position regularized support vector domain
  description},
\newblock \bibinfo{journal}{Pattern Recognition} \bibinfo{volume}{46 (3)}
  (\bibinfo{year}{2013}) \bibinfo{pages}{875--884}.
\bibitem[{Jain(2010)}]{jain10_survey}
\bibinfo{author}{A.~K. Jain},
\newblock \bibinfo{title}{Data clustering: 50 years beyond $k$-means},
\newblock \bibinfo{journal}{Pattern Recognition Letters} \bibinfo{volume}{31
  (8)} (\bibinfo{year}{2010}) \bibinfo{pages}{651--666}.
\bibitem[{Vega-Pons and Ruiz-Shulcloper(2011)}]{vega_pons11_survey}
\bibinfo{author}{S.~Vega-Pons}, \bibinfo{author}{J.~Ruiz-Shulcloper},
\newblock \bibinfo{title}{A survey of clustering ensemble algorithms},
\newblock \bibinfo{journal}{International Journal of Pattern Recognition and
  Artificial Intelligence} \bibinfo{volume}{25 (3)} (\bibinfo{year}{2011})
  \bibinfo{pages}{337--372}.
\bibitem[{Strehl and Ghosh(2003)}]{strehl02}
\bibinfo{author}{A.~Strehl}, \bibinfo{author}{J.~Ghosh},
\newblock \bibinfo{title}{Cluster ensembles: A knowledge reuse framework for
  combining multiple partitions},
\newblock \bibinfo{journal}{Journal of Machine Learning Research}
  \bibinfo{volume}{3} (\bibinfo{year}{2003}) \bibinfo{pages}{583--617}.
\bibitem[{Fern and Brodley(2004)}]{fern04_bipartite}
\bibinfo{author}{X.~Z. Fern}, \bibinfo{author}{C.~E. Brodley},
\newblock \bibinfo{title}{Solving cluster ensemble problems by bipartite graph
  partitioning},
\newblock in: \bibinfo{booktitle}{Proceedings of the International Conference
  on Machine Learning (ICML'04), 2004}.
\bibitem[{Fred and Jain(2005)}]{Fred05_EAC}
\bibinfo{author}{A.~L.~N. Fred}, \bibinfo{author}{A.~K. Jain},
\newblock \bibinfo{title}{Combining multiple clusterings using evidence
  accumulation},
\newblock \bibinfo{journal}{IEEE Transactions on Pattern Analysis and Machine
  Intelligence} \bibinfo{volume}{27 (6)} (\bibinfo{year}{2005})
  \bibinfo{pages}{835--850}.
\bibitem[{Topchy et~al.(2005)Topchy, Jain, and Punch}]{topchy05}
\bibinfo{author}{A.~Topchy}, \bibinfo{author}{A.~K. Jain},
  \bibinfo{author}{W.~Punch},
\newblock \bibinfo{title}{Clustering ensembles: models of consensus and weak
  partitions},
\newblock \bibinfo{journal}{IEEE Transactions on Pattern Analysis and Machine
  Intelligence} \bibinfo{volume}{27 (12)} (\bibinfo{year}{2005})
  \bibinfo{pages}{1866--1881}.
\bibitem[{Hadjitodorov et~al.(2006)Hadjitodorov, Kuncheva, and
  Todorova}]{diversity06}
\bibinfo{author}{S.~T. Hadjitodorov}, \bibinfo{author}{L.~I. Kuncheva},
  \bibinfo{author}{L.~P. Todorova},
\newblock \bibinfo{title}{Moderate diversity for better cluster ensembles},
\newblock \bibinfo{journal}{Information Fusion} \bibinfo{volume}{7 (3)}
  (\bibinfo{year}{2006}) \bibinfo{pages}{264--275}.
\bibitem[{Li et~al.(2007)Li, Yu, Hao, and Li}]{li07}
\bibinfo{author}{Y.~Li}, \bibinfo{author}{J.~Yu}, \bibinfo{author}{P.~Hao},
  \bibinfo{author}{Z.~Li},
\newblock \bibinfo{title}{Clustering ensembles based on normalized edges},
\newblock in: \bibinfo{booktitle}{Proceedings of the Pacific-Asia Conference on
  Knowledge Discovery and Data Mining (PAKDD'07), 2007}.
\bibitem[{Iam-On et~al.(2008)Iam-On, Boongoen, and Garrett}]{iamon08_icds}
\bibinfo{author}{N.~Iam-On}, \bibinfo{author}{T.~Boongoen},
  \bibinfo{author}{S.~Garrett},
\newblock \bibinfo{title}{Refining pairwise similarity matrix for cluster
  ensemble problem with cluster relations},
\newblock in: \bibinfo{booktitle}{Proceedings of the International Conference
  on Discovery Science (ICDS'08), 2008}.
\bibitem[{Domeniconi and Al-Razgan(2009)}]{Domeniconi09}
\bibinfo{author}{C.~Domeniconi}, \bibinfo{author}{M.~Al-Razgan},
\newblock \bibinfo{title}{Weighted cluster ensembles: Methods and analysis},
\newblock \bibinfo{journal}{ACM Transactions on Knowledge Discovery from Data}
  \bibinfo{volume}{2 (4)} (\bibinfo{year}{2009}) \bibinfo{pages}{1--40}.
\bibitem[{Wang et~al.(2009)Wang, Yang, and Zhou}]{wang09_pr}
\bibinfo{author}{X.~Wang}, \bibinfo{author}{C.~Yang},
  \bibinfo{author}{J.~Zhou},
\newblock \bibinfo{title}{Clustering aggregation by probability accumulation},
\newblock \bibinfo{journal}{Pattern Recognition} \bibinfo{volume}{42 (5)}
  (\bibinfo{year}{2009}) \bibinfo{pages}{668--675}.
\bibitem[{Mimaroglu and Erdil(2011)}]{Mimaroglu11_pr}
\bibinfo{author}{S.~Mimaroglu}, \bibinfo{author}{E.~Erdil},
\newblock \bibinfo{title}{Combining multiple clusterings using similarity
  graph},
\newblock \bibinfo{journal}{Pattern Recognition} \bibinfo{volume}{44 (3)}
  (\bibinfo{year}{2011}) \bibinfo{pages}{694--703}.
\bibitem[{Iam-On et~al.(2011)Iam-On, Boongoen, Garrett, and
  Price}]{iam_on11_linkbased}
\bibinfo{author}{N.~Iam-On}, \bibinfo{author}{T.~Boongoen},
  \bibinfo{author}{S.~Garrett}, \bibinfo{author}{C.~Price},
\newblock \bibinfo{title}{A link-based approach to the cluster ensemble
  problem},
\newblock \bibinfo{journal}{IEEE Transactions on Pattern Analysis and Machine
  Intelligence} \bibinfo{volume}{33 (12)} (\bibinfo{year}{2011})
  \bibinfo{pages}{2396--2409}.
\bibitem[{Yi et~al.(2012)Yi, Yang, Jin, and Jain}]{yi_icdm12}
\bibinfo{author}{J.~Yi}, \bibinfo{author}{T.~Yang}, \bibinfo{author}{R.~Jin},
  \bibinfo{author}{A.~K. Jain},
\newblock \bibinfo{title}{Robust ensemble clustering by matrix completion},
\newblock in: \bibinfo{booktitle}{Proceedings of the IEEE International
  Conference on Data Mining (ICDM'12), 2012}.
\bibitem[{Franek and Jiang(2014)}]{franek13_pr}
\bibinfo{author}{L.~Franek}, \bibinfo{author}{X.~Jiang},
\newblock \bibinfo{title}{Ensemble clustering by means of clustering embedding
  in vector spaces},
\newblock \bibinfo{journal}{Pattern Recognition} \bibinfo{volume}{47 (2)}
  (\bibinfo{year}{2014}) \bibinfo{pages}{833--842}.
\bibitem[{Huang et~al.(2013)Huang, Lai, and Wang}]{huang_iscide13}
\bibinfo{author}{D.~Huang}, \bibinfo{author}{J.-H. Lai}, \bibinfo{author}{C.-D.
  Wang},
\newblock \bibinfo{title}{Exploiting the wisdom of crowd: A multi-granularity
  approach to clustering ensemble},
\newblock in: \bibinfo{booktitle}{Proceedings of the International Conference
  on Intelligence Science and Big Data Engineering (IScIDE'13), 2013}.
\bibitem[{Cristofor and Simovici(2002)}]{cristofor02}
\bibinfo{author}{D.~Cristofor}, \bibinfo{author}{D.~Simovici},
\newblock \bibinfo{title}{Finding median partitions using
  information-theoretical-based genetic algorithms},
\newblock \bibinfo{journal}{Journal of Universal Computer Science}
  \bibinfo{volume}{8 (2)} (\bibinfo{year}{2002}) \bibinfo{pages}{153--172}.
\bibitem[{Weiszfeld and Plastria(2009)}]{Weiszfeld09}
\bibinfo{author}{E.~Weiszfeld}, \bibinfo{author}{F.~Plastria},
\newblock \bibinfo{title}{On the point for which the sum of the distances to n
  given points is minimum},
\newblock \bibinfo{journal}{Annals of Operations Research} \bibinfo{volume}{167
  (1)} (\bibinfo{year}{2009}) \bibinfo{pages}{7--41}.
\bibitem[{Vega-Pons et~al.(2010)Vega-Pons, Correa-Morris, and
  Ruiz-Shulcloper}]{vega_pons10}
\bibinfo{author}{S.~Vega-Pons}, \bibinfo{author}{J.~Correa-Morris},
  \bibinfo{author}{J.~Ruiz-Shulcloper},
\newblock \bibinfo{title}{Weighted partition consensus via kernels},
\newblock \bibinfo{journal}{Pattern Recognition} \bibinfo{volume}{43 (8)}
  (\bibinfo{year}{2010}) \bibinfo{pages}{2712--2724}.
\bibitem[{Vega-Pons et~al.(2011)Vega-Pons, Ruiz-Shulcloper, and
  Guerra-Gand\'{o}n}]{vega_pons_PRL11}
\bibinfo{author}{S.~Vega-Pons}, \bibinfo{author}{J.~Ruiz-Shulcloper},
  \bibinfo{author}{A.~Guerra-Gand\'{o}n},
\newblock \bibinfo{title}{Weighted association based methods for the
  combination of heterogeneous partitions},
\newblock \bibinfo{journal}{Pattern Recognition Letters} \bibinfo{volume}{32
  (16)} (\bibinfo{year}{2011}) \bibinfo{pages}{2163--2170}.
\bibitem[{Li and Ding(2008)}]{Li_WCC08}
\bibinfo{author}{T.~Li}, \bibinfo{author}{C.~Ding},
\newblock \bibinfo{title}{Weighted consensus clustering},
\newblock in: \bibinfo{booktitle}{Proceedings of the SIAM International
  Conference on Data Mining (SDM'08), 2008}.
\bibitem[{Fern and Lin(2008)}]{Fern08_selection}
\bibinfo{author}{X.~Z. Fern}, \bibinfo{author}{W.~Lin},
\newblock \bibinfo{title}{Cluster ensemble selection},
\newblock \bibinfo{journal}{Statistical Analysis and Data Mining}
  \bibinfo{volume}{1 (3)} (\bibinfo{year}{2008}) \bibinfo{pages}{128--141}.
\bibitem[{Wu and Chow(2004)}]{wu04_validity}
\bibinfo{author}{S.~Wu}, \bibinfo{author}{T.~W.~S. Chow},
\newblock \bibinfo{title}{Clustering of the self-organizing map using a
  clustering validity index based on inter-cluster and intra-cluster density},
\newblock \bibinfo{journal}{Pattern Recognition} \bibinfo{volume}{37 (2)}
  (\bibinfo{year}{2004}) \bibinfo{pages}{175--188}.
\bibitem[{Faceli et~al.(2009)Faceli, de~Souto, de~Ara\'{u}jo, and
  de~Carvalho}]{faceli09}
\bibinfo{author}{K.~Faceli}, \bibinfo{author}{M.~C.~P. de~Souto},
  \bibinfo{author}{D.~S.~A. de~Ara\'{u}jo}, \bibinfo{author}{A.~C. P. L.~F.
  de~Carvalho},
\newblock \bibinfo{title}{Multi-objective clustering ensemble for gene
  expression data analysis},
\newblock \bibinfo{journal}{Neurocomputing} \bibinfo{volume}{72}
  (\bibinfo{year}{2009}) \bibinfo{pages}{2763--2774}.
\bibitem[{Li and Latecki(2012)}]{li12_nips}
\bibinfo{author}{N.~Li}, \bibinfo{author}{L.~J. Latecki},
\newblock \bibinfo{title}{Clustering aggregation as maximum-weight independent
  set},
\newblock in: \bibinfo{booktitle}{Advances in Neural Information Processing
  Systems (NIPS'12), 2012}.
\bibitem[{Surowiecki(2004)}]{Surowiecki04}
\bibinfo{author}{J.~Surowiecki}, \bibinfo{title}{The wisdom of crowds: Why the
  many are smarter than the few and how collective wisdom shapes business,
  economies, societies and nations}, \bibinfo{publisher}{Anchor Books},
  \bibinfo{year}{2004}.
\bibitem[{Levandowsky and Winter(1971)}]{levan71_nature}
\bibinfo{author}{M.~Levandowsky}, \bibinfo{author}{D.~Winter},
\newblock \bibinfo{title}{Distance between sets},
\newblock \bibinfo{journal}{Nature} \bibinfo{volume}{234}
  (\bibinfo{year}{1971}) \bibinfo{pages}{34--35}.
\bibitem[{Li et~al.(2012)Li, Wu, and Chang}]{CVPR12_Li}
\bibinfo{author}{Z.~Li}, \bibinfo{author}{X.-M. Wu}, \bibinfo{author}{S.-F.
  Chang},
\newblock \bibinfo{title}{Segmentation using superpixels: A bipartite graph
  partitioning approach},
\newblock in: \bibinfo{booktitle}{Proceedings of the IEEE Conference on
  Computer Vision and Pattern Recognition (CVPR'12), 2012}.
\bibitem[{Bache and Lichman(2013)}]{Bache+Lichman:2013}
\bibinfo{author}{K.~Bache}, \bibinfo{author}{M.~Lichman}, \bibinfo{title}{{UCI}
  machine learning repository}, \bibinfo{year}{2013}.
\bibitem[{Huang et~al.(2012)Huang, Lai, and Wang}]{huang2012incremental}
\bibinfo{author}{D.~Huang}, \bibinfo{author}{J.-H. Lai}, \bibinfo{author}{C.-D.
  Wang},
\newblock \bibinfo{title}{Incremental support vector clustering with outlier
  detection},
\newblock in: \bibinfo{booktitle}{Proceedings of the International Conference
  on Pattern Recognition (ICPR'12), 2012}.

\end{thebibliography}

\end{document}